%% file: main.tex
\documentclass[10pt,twocolumn,letterpaper]{article}

\usepackage{cvpr}      

\usepackage{graphicx}
\usepackage{amsmath}
\usepackage{amssymb}
\usepackage{booktabs}
\usepackage[linesnumbered,ruled]{algorithm2e}
\usepackage{multicol}
\usepackage{multirow}
\usepackage[accsupp]{axessibility}

\usepackage[pagebackref,breaklinks,colorlinks]{hyperref}

\usepackage[capitalize]{cleveref}
\crefname{section}{Sec.}{Secs.}
\Crefname{section}{Section}{Sections}
\Crefname{table}{Table}{Tables}
\crefname{table}{Tab.}{Tabs.}

\def\cvprPaperID{15} 
\def\confName{CVPR}
\def\confYear{2023}

\begin{document}

\title{TimelyFL: Heterogeneity-aware Asynchronous Federated Learning with \\
Adaptive Partial Training}

\author{Tuo Zhang$^1$, Lei Gao$^1$, Sunwoo Lee$^2$, Mi Zhang$^3$, Salman Avestimehr$^1$\\
$^1$University of Southern California \\$^2$Inha University\\ $^3$The Ohio State University\\
{\tt\small tuozhang@usc.edu}}

\maketitle

\begin{abstract}
In cross-device Federated Learning (FL) environments, scaling synchronous FL methods is challenging as stragglers hinder the training process. Moreover, the availability of each client to join the training is highly variable over time due to system heterogeneities and intermittent connectivity. Recent asynchronous FL methods (e.g., \texttt{FedBuff} \cite{Nguyen2021FederatedLW}) have been proposed to overcome these issues by allowing slower users to continue their work on local training based on stale models and to contribute to aggregation when ready. However, we show empirically that this method can lead to a substantial drop in training accuracy as well as a slower convergence rate. The primary reason is that fast-speed devices contribute to many more rounds of aggregation while others join more intermittently or not at all, and with stale model updates. To overcome this barrier, we propose \texttt{TimelyFL}, a heterogeneity-aware asynchronous FL framework with adaptive partial training. During the training, \texttt{TimelyFL} adjusts the local training workload based on the real-time resource capabilities of each client, aiming to allow more available clients to join in the global update without staleness. We demonstrate the performance benefits of  \texttt{TimelyFL} by conducting extensive experiments on various datasets (e.g., CIFAR-10, Google Speech, and Reddit) and models (e.g., ResNet20, VGG11, and ALBERT). In comparison with the state-of-the-art (i.e., \texttt{FedBuff}), our evaluations reveal that \texttt{TimelyFL} improves participation rate by 21.13\%, harvests 1.28$\times$ - 2.89$\times$ more efficiency on convergence rate, and provides a 6.25\% increment on test accuracy.
\end{abstract}

\input{sections/1_intro}
\input{sections/2_related}
\input{sections/3_design}
\input{sections/4_exp}

\input{sections/5_conclusion}

{\small
\bibliographystyle{ieee_fullname}
\bibliography{egbib}
}
\input{sections/appendix}
\end{document}

%% file: sections/1_intro.tex
\section{Introduction}
Federated learning (FL) has emerged as a promising distributed machine learning paradigm that preserves privacy~\cite{kairouz2021advances,wang2021field}. The gist of FL is to keep the clients' private data on the devices and perform local model training for each client. A central server will collect these locally trained models to update a global model and then push it back for the next round of training.

Most existing FL protocols are based on synchronous FL training (SyncFL), meaning that at each round all clients (or a selected cohort of clients) are updating their local models based on the latest update broadcast by the server at the beginning of that round. Due to the unbalanced communication or hardware capabilities and non-identical training data distribution, however, the time consumption for a local update can vary substantially from device to device, and some clients may even be temporarily disconnected during the training process~\cite{circul}. Thus, leaving the server with two suboptimal choices: to wait for \emph{all} clients participating in each round to finish their local training and contribute to model aggregation (which will cause significant delays due to stragglers), or to only wait for a \emph{subset} of the faster clients (which will ignore all the work and contributions from slower clients). These critical challenges largely impede the scalability of SyncFL and make it difficult to land in large-scale cross-device scenarios.

To address those challenges, recent works have proposed asynchronous federated learning (AsyncFL) \cite{Huba2021PapayaPP, Nguyen2021FederatedLW, Wu2021SAFAAS, Avdiukhin2021FederatedLU}, which allows slower clients to continue local training and contribute to future aggregation rounds.
AsyncFL \textit{decouples} client local training from global model aggregation/updates, as only certain clients would simultaneously get an update from the cloud server, which decreases the impact of stragglers. The most recent AsyncFL work – \texttt{FedBuff}~\cite{Nguyen2021FederatedLW, Huba2021PapayaPP} – proposes that the server should perform a gradient aggregation to create a global model once the number of received local updates reaches a requisite threshold, which is a tunable parameter referred as \textit{aggregation goal}. The slower clients can still upload their updates later as long as they finish local training, but their updates may not be included based on staleness information. 

\begin{figure*}[h!]
     \centering
     \begin{subfigure}[b]{0.66\columnwidth}
         \centering
         \includegraphics[width=\columnwidth]{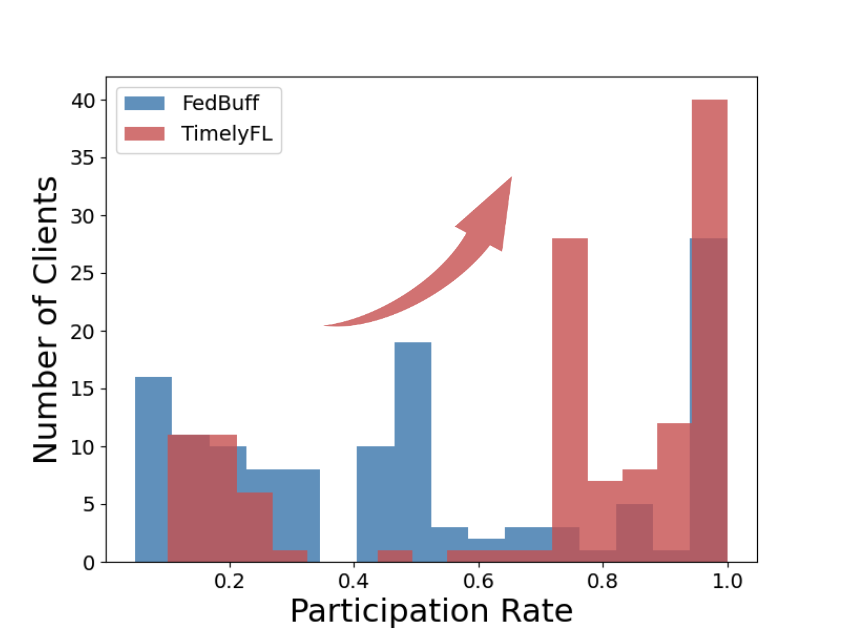}
         \caption{Participation rate distribution across all devices}
         \label{fig:1a}
     \end{subfigure}
     \hfill
     \begin{subfigure}[b]{0.66\columnwidth}
         \centering
         \includegraphics[width=\columnwidth]{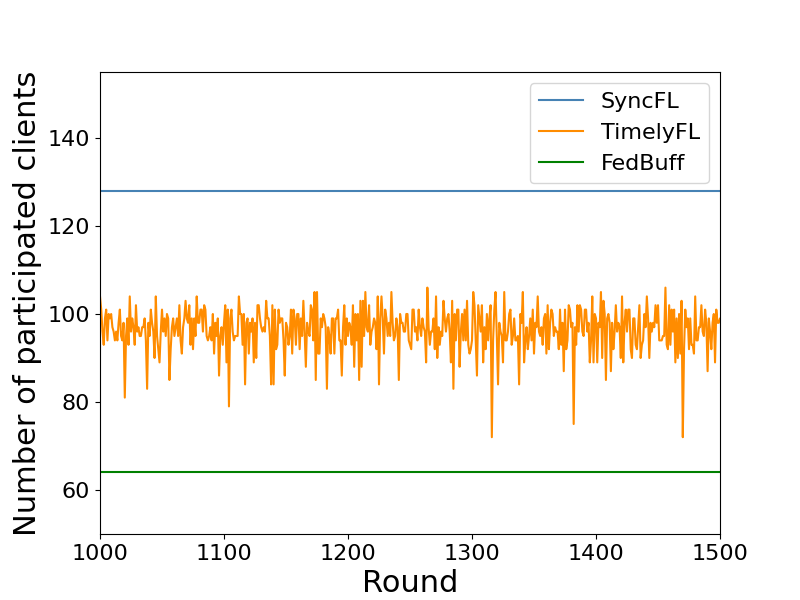}
         \caption{Number of participated clients during FL training rounds}
         \label{fig:1b}
     \end{subfigure}
     \hfill
     \begin{subfigure}[b]{0.66\columnwidth}
         \centering
         \includegraphics[width=\columnwidth]{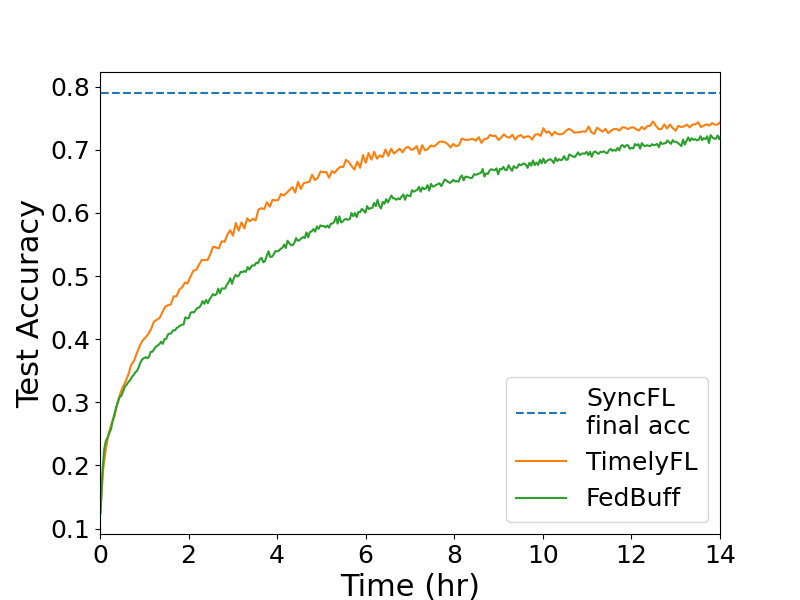}
         \caption{Time-to-accuracy performance for each strategy}
         \label{fig:1c}
     \end{subfigure}
        \caption{Empirical performance of SyncFL, \texttt{FedBuff}, and \texttt{TimelyFL} in CIFAR-10 classification task with FedOpt as server aggregator (for experiment details and other evaluations see Section~\ref{experiment}). \texttt{TimelyFL} includes more devices join in global update during the training (shown in (a) and (b)). As more devices participate timely, \texttt{TimelyFL} harvests both convergence rate and accuracy boost compared to \texttt{FedBuff} (shown in (c)).}
        \label{fig:1}
\end{figure*}

As highlighted in Figure~\ref{fig:1c}, we empirically demonstrate that while \texttt{FedBuff} achieves much faster convergence to a certain intermediate accuracy, it can, unfortunately, lead to a substantial drop in final accuracy compared to SyncFL. The intuitive explanation is that, as \texttt{FedBuff} only accepts a \textit{fixed} number of local updates to contribute to the global model in every communication round, it decreases the parallel computing efficiency by blocking other completed local updates into global aggregation, which turns them into stale updates as they would be postponed to the next round of global update. Moreover, the server aggregator prefers the fast-speed devices contributing more rounds of training, whereas low-speed devices may not enjoy the same frequency of contribution. Even when the slow devices participate in global training, they occasionally send the staled updates that potentially harm the convergence of the global loss. As shown in Figure~\ref{fig:1a} and \ref{fig:1b}, compared to SyncFL, \texttt{FedBuff} only includes a fixed number of local updates per round, and achieves a low participation rate (i.e., the number of aggregation participated divided by the total number of aggregation rounds) on average with a biased distribution, indicating that the inclusiveness of a group has been diminished, which is the root cause of the test accuracy gap.

To close the gap between SyncFL and AsyncFL, we propose \texttt{TimelyFL}, a heterogeneity-aware asynchronous federated learning framework based on adaptive partial training.
One key distinction of \texttt{TimelyFL} from previous AsyncFL works is that \texttt{TimelyFL} does not set a fixed number limit to the number of local updates for the global aggregation per round. Instead, to accommodate a \textit{flexible} number of clients joining in the global update, we set a wall-clock time limit for every aggregation interval. The aggregation interval equals the $k$th fastest local update time among all clients, where $k$ is a tunable parameter. As long as the device can deliver its model update to the server within this interval, it will be part of the global aggregation. To include more available devices to join in global aggregation without staleness, we introduce \textit{partial model training} for clients with low capacity. Instead of fully training a model, only a part of the model that composes of a subset of consecutive output-side layers will be assigned to them for backward pass training. With partial model training, both local computation time and communication time will decrease for stale clients.

\begin{figure*}
  \hspace*{0.2in}
  \includegraphics[width=1.8\columnwidth, height = 6.3cm]{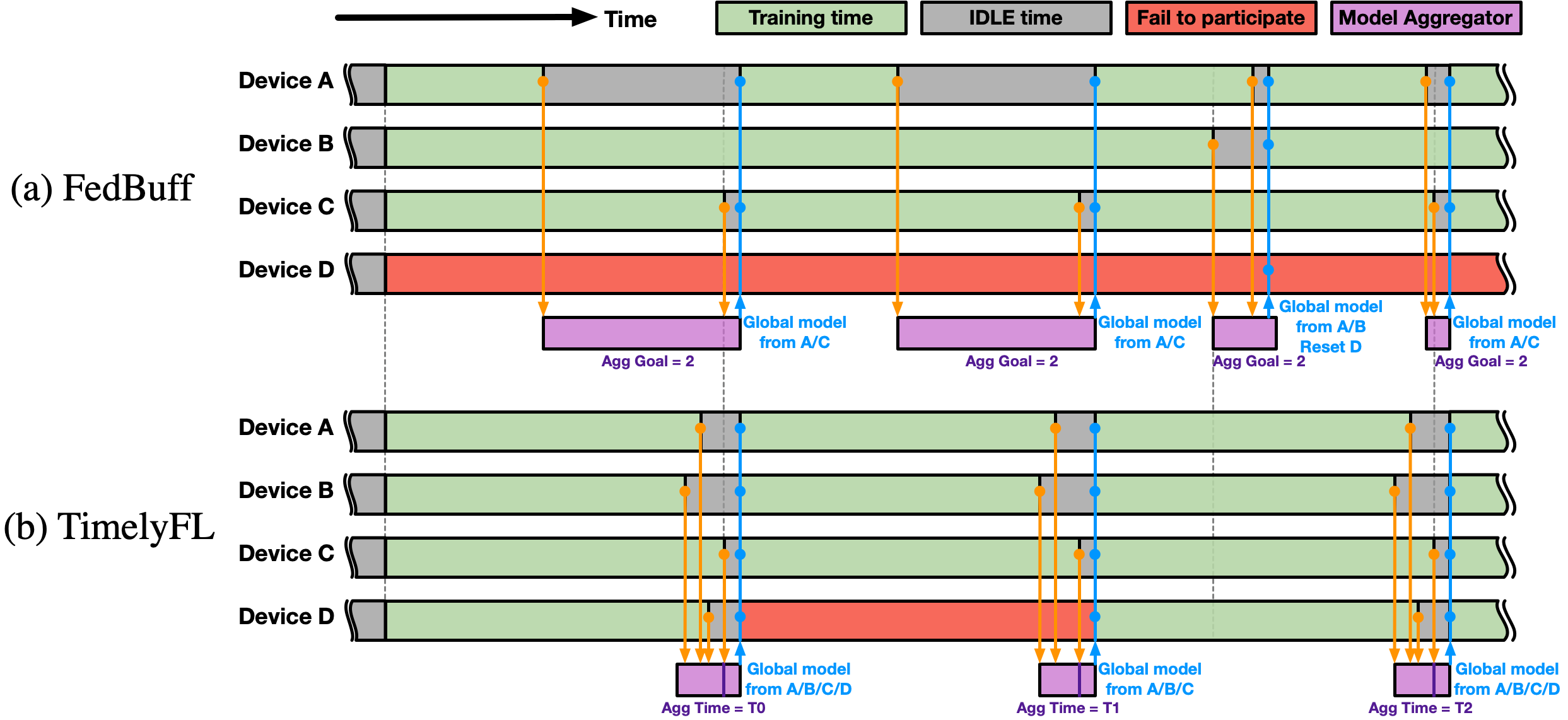}
  \caption{\texttt{FedBuff} (top): Server updates the global model as it receives the requisite number of local updates, and slower devices still could send their updates at a later time to the server. Fast devices participate more times in the global update, while slow devices contribute less or no participation.
  \texttt{TimelyFL} (bottom): Server updates the individual workload every round based on the real-time availability of each client to include more devices in global update timely, largely increases the participation rate for slow devices.}
  \label{fig:time}
\end{figure*}

As shown in Figure~\ref{fig:time}, \texttt{TimelyFL} unifies the local training time by adaptively adjusting the workload (i.e., the local epoch number and partial training ratio) for each client, making it feasible for clients to finish the local training and upload the updates to the server within the calculated aggregation interval in every communication round. As such, \texttt{TimelyFL} tackles the system heterogeneity issue and eliminates the staleness of local update reports for slower devices.

We evaluate the performance of \texttt{TimelyFL} across various application tasks, including image classification, speech recognition, and natural language processing on CIFAR-10~\cite{Krizhevsky2009LearningML}, Google Speech Command~\cite{Warden2018SpeechCA}, and Reddit Comment~\cite{reddit} benchmark datasets, respectively, with two commonly used aggregation functions, FedAvg \cite{McMahan2017CommunicationEfficientLO} and FedOpt \cite{Reddi2021AdaptiveFO}. Our results show that 66.4\% of devices increase the participation rate and the average participation rate increases by 21.1\% in \texttt{TimelyFL} compared to \texttt{FedBuff}. Under the same scale of the FL system, \texttt{TimelyFL} outperforms \texttt{FedBuff} \cite{Nguyen2021FederatedLW} on both time-to-accuracy and final test accuracy, as shown in Figure~\ref{fig:1c}. 

%% file: sections/2_related.tex
\section {Related Work} \label{related-work}

\textbf{Asynchronous Federated Learning.}
Due to intermittent connectivity and availability among clients, asynchronous FL is a promising solution to tackle device  heterogeneity in FL \cite{fliotvision2022ieeeiotm}. Most  asynchronous FL works concentrate on solving the straggler problem, such as \cite{Li2021StragglersAN}, \cite{Wu2021SAFAAS}, and \cite{Xie2019AsynchronousFO}. 
\texttt{PAPAYA} \cite{Huba2021PapayaPP} and \texttt{FedBuff} \cite{Nguyen2021FederatedLW} have been proposed to mitigate stragglers and enable secure aggregation jointly. Specifically, the individual updates are not incorporated by the server as soon they arrive. Instead, the server will keep receiving local updates in a secure buffer, where the buffer size is a tunable parameter, and update the global model whenever the buffer is full. The slow devices can also send the local update to the server after the global aggregation. Their update will be considered for the next available global update. However, practically speaking, fast devices participate in global updates many more times than slow devices, and some slow devices cannot join in the global aggregation even once due to the staleness control.

All of the above approaches assume that the client should process the local training within the full-model size. As the slower users participate in the global aggregation, they can only contribute with stale updates. Some previous works have pointed out that the effects of the stale update on distributed machine learning can directly harm the overall convergence behavior for the final model, aligned with the asynchronous distributed optimization theory suggested by~\cite{Zhou2018DistributedAO, Dai2019TowardUT, Giladi2020AtSE}. Moreover, the participation rate is mainly unbalanced due to the high-speed devices contributing more rounds to global updates compared to the slow-speed devices. In contrast to previous approaches, we focus on enabling all clients to join in the global aggregation effectively based on their local resources to improve the inclusiveness of the final global model training. 

\textbf{Partial Model Training.} Partial model training can be viewed as an efficient approach to reduce both communication and computation workload on the client-side of the FL system \cite{fedrolex2022neurips}. \texttt{FedPrune} \cite{Munir2021FedPruneTI} proposes a method that prunes the global model for each client based on their device capabilities, where slow clients are served smaller models and
faster clients train on larger models. \texttt{FedPT} \cite{Sidahmed2021EfficientAP} leverages the partially trainable neural networks on clients to reduce communication costs and enable faster training with a smaller memory footprint and with few implications on model performance. Other works such as \cite{Yang2021PartialVT, Ro2022ScalingLM, Lee2021LayerwiseAM} also address that partial model training can save both communication cost and memory usage in cross-device FL. All of the above works
maintain the partial ratio for the sub-model of a certain client as constant 
during the entire FL training process, which neglects that the availability of each device is not stable throughout the time. In this work, we adaptively adjust the partial ratio for the local model training based on the real-time device availability, which aims to improve both efficiency and utility for each client.

%% file: sections/3_design.tex
\begin{figure*}[t]
\centering
\includegraphics[width=1.8\columnwidth, height = 5cm]{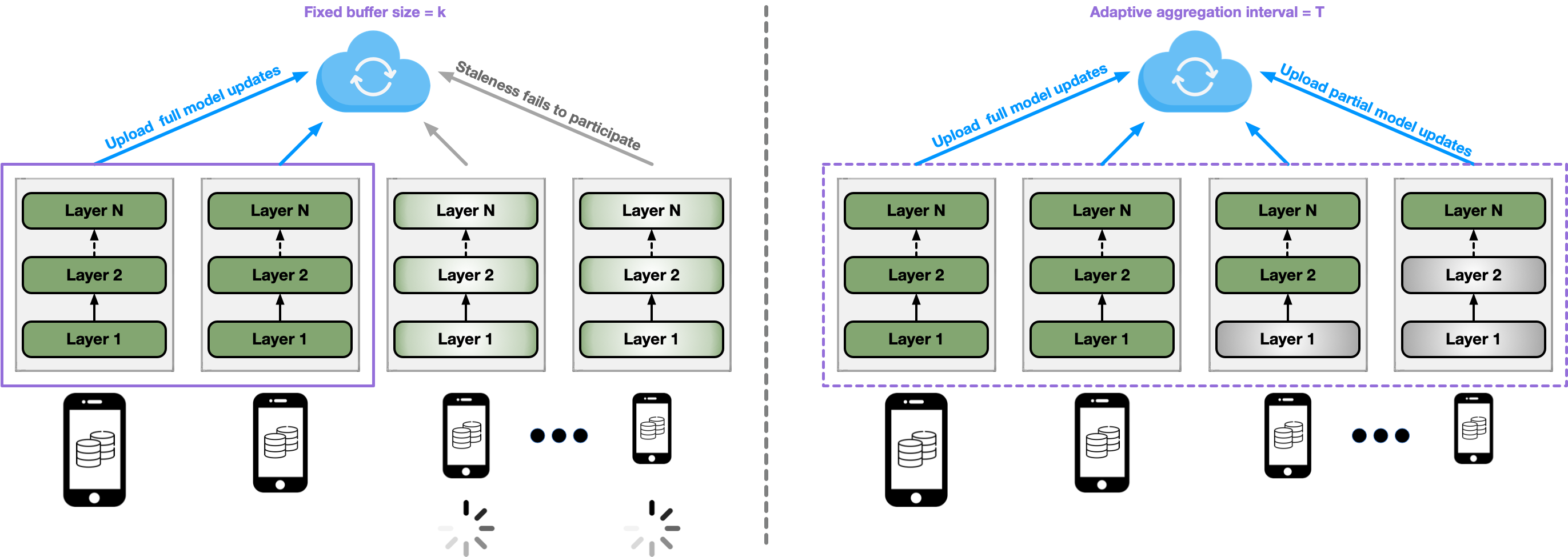}
\caption{
\textbf{Left}: The standard AsyncFL framework. The server will initiate the global update once it collects the requisite number of local updates. The other clients will be postponed to a latter communication round with stale update information.
\textbf{Right}: The proposed \texttt{TimelyFL}. The server will include all the received local updates within aggregation interval to global update. Clients with a weaker capacity are assigned to train a subset of the model to catch the aggregation interval time.
}
\label{fig:partial}
\end{figure*}

\section {Our Method} \label{design}

\subsection{Standard Asynchronous Federated Learning}

Figure~\ref{fig:partial} (left) illustrates the standard asynchronous FL framework. Instead of waiting for all clients to finish the local model training, the server stores the individual updates from clients in a buffer and then adjusts the global model once the buffer size reaches the requisite number of the aggregation goal. Other non-participating devices will postpone their contribution to global updates in the latter communication round once they finish the training. Given that the standard AsyncFL framework suffers from inclusiveness constraints described in the introduction section, we propose an efficient AsyncFL framework as shown in Figure~\ref{fig:partial} (right) to address this issue.

\subsection{TimelyFL Design}

\subsubsection{Preliminaries}
To increase the participation rate for the slow clients, we aim to design a cross-device asynchronous federated learning framework where each device can finish the local update within a limited time interval by adaptively adjusting its workload per round. Here, the workload is defined as the product of local training epoch number $E$ and partial model training ratio $\alpha$. To formalize this, our time utility function, which considers the local training optimization at the client side, is designed as follows:
\begin{equation}
    \underset{{E}, \alpha}{\arg\max}~~~(\Tilde{t}_{cmp, c} \times E \times \alpha + \Tilde{t}_{com, c} \times \alpha) \leq {T_k}
    \label{eq1}
\end{equation}
where $\Tilde{t}_{cmp, c}$ is the estimated local computation time, and $\Tilde{t}_{com, c}$ is the estimated local communication time of the client $c$ for one epoch of full model training in a certain communication round calculated by the server. Note that both $\Tilde{t}_{cmp, c}$ and $\Tilde{t}_{com, c}$ are not constant throughout the training due to the nature of mobile devices.
By adjusting $E$ and $\alpha$, each device is supposed to participate in the global aggregation every communication round timely and maximally utilize its resource capability within aggregation interval $T_k$.
Therefore, the overall distributed optimization involves more iterations on diverse datasets, leading to faster convergence.

\subsubsection{Adaptive Model Training}

Due to resource limitations, some weak devices may not finish the full model training effectively within the time interval $T_k$, making them become stale clients in the system and impeding them from contributing to the global model. To address this issue, we introduce partial model training to slow devices. Figure~\ref{fig:partial} (right) illustrates our approach when partial model training lands on the device heterogeneity FL system. Instead of a full training model, weak devices will be assigned to train partial models composed of a subset of consecutive output-side layers. During the training, only the subset of assigned layers will operate both forward pass and backward propagation, while the other layers will only process the forward pass for the input data but be frozen during weight updates. After local training finishes, the partially-trained clients only return the update for the assigned layers to the server for the global aggregation, as the frozen layers are unchanged during training.

We target to solve two bottlenecks in the cross-device FL with partial model training, communication and computation efficiency for the stale clients. In edge computing, the backward propagation consumes much more time than the forward pass. Partial model training would essentially reduce the training time, as it does not need to calculate gradients for the frozen parameters. The reduced time is roughly proportional to the reduced model size, as we empirically shown in the Appendix~\ref{partial_experiment}. Moreover, we only send the trainable part of updates to the server, substantially improving communication efficiency, especially when stragglers with limited network connections exist. By implementing partial model training, we aim to let low-capacity devices report their local updates to the server timely without staleness, thereby improving their participation rate during FL training. 

\subsubsection{TimelyFL Algorithm}
Based on the adaptive model training, we propose the \texttt{TimelyFL}.
\texttt{TimelyFL} tries to unify each client's round time to the limited aggregation interval $T_k$ by adaptively adjusting the workload concerning its real-time availability per communication round. The workload is defined as the product of the partial training ratio $\alpha$ and the local epoch number $E$. \texttt{TimelyFL} framework is composed of three main parts, \texttt{TimelyFL} server, local time update, and workload scheduling.

\begin{algorithm}[h!]
\SetKwInput{Input}{Input}
\SetKwInput{Output}{Output}
    \Input{$k$: the aggregation participation target, $n$: the number of training concurrency}
    \For{$r \in \{0, \cdots, R-1 \}$ \textbf{communication rounds}}{
         {\textbf{Global server do:}}\\
            \hspace*{1em} Sample $n$ clients uniformly at random to define $\mathcal{S}$, and send ${W}^{r}_{s}$ to clients in $\mathcal{S}$ \;
         {\textbf{Clients $c \in\mathcal{S}$ in parallel do:}}\\
            \hspace*{1em} $\Tilde{t}_{total}, \Tilde{t}_{cmp}, \Tilde{t}_{com} = \textrm{LocalTimeUpdate}(M)$ \;
         {\textbf{Global server do:}}\\
            \hspace*{1em} $T_k^r \leftarrow$ the $k$th smallest number in $\langle{\Tilde{t}_{total}}\rangle$ \;
            \hspace*{1em} $\langle{{E}^{r}}\rangle, \langle{\alpha^{r}}\rangle, \langle{t_{rpt}^{r}}\rangle = \textrm{WorkloadScheduling}({T_k}^r, \langle{\Tilde{t}_{cmp}}\rangle, \langle{\Tilde{t}_{com}}\rangle)$ \;
         {\textbf{Clients $c \in\mathcal{S}$ in parallel do:}}\\
            \hspace*{1em} ${W}^{r}_{c}$ $\leftarrow$ adaptive model training \;
         {\textbf{Global server do:}}\\
            \hspace*{1em} ${W}^{r+1}_{s}$ $\leftarrow$ aggregate $\langle{{W}^{r}_{c}}\rangle$ \;
        }
    \Output{${W}^{R}_{s}$}
\caption{TimelyFL.}
\label{alg:dynamicfl}
\end{algorithm}

\textbf{\texttt{TimelyFL} Server.} \texttt{TimelyFL} server is in charge of adjusting the aggregation interval $T_k$, local training epoch $E$, and partial training ratio $\alpha$ for each device during the FL training, as summarized in Algorithm~\ref{alg:dynamicfl}. The aggregation interval $T_k$ in each round equals the $k$th smallest value among $\langle{\Tilde{t}_{total}}\rangle$, as the estimated unit total time for all clients. 
At each communication round, \texttt{TimelyFL} server randomly samples $n$ clients to construct the collection $\mathcal{S}$ and distributes the global model to the clients inside $\mathcal{S}$, which means $n$ clients would start the local training in this round, same as the definition of training concurrency in the \texttt{FedBuff}. Each selected client would perform one data batch full model training to estimate its time consumption and report it to the server. Then, aggregation interval time $T_k$ and training hyperparameters for client $c$ (i.e., local training epoch number $E$ and partial training ratio $\alpha$) would be adjusted based on all selected clients' status during the FL training process. The server would also return a local computation budget time $t_{rpt,c}$, as the wall-clock time when the client must report its training status. 


\textbf{Local Time Update.} To efficiently accommodate the capabilities, each participant needs to update its time consumption to the server as summarized in Algorithm~\ref{alg:timeupdate}. Specifically, each client would collect the real computation time $t_{cmp}$ from one data batch full model training. The unit computation time $\Tilde{t}_{cmp}$ is estimated by $t_{cmp}$ and progress $\beta$, where $\beta$ is defined as the ratio of trained batch number to the total data batch number. The local communication time equals the model's file size $M$ over the device's real-time network bandwidth $Bw$, as the same setting in the previous FL system work \cite{Lai2021FedScaleBM}.

\begin{algorithm}[h]
\SetKwInput{Input}{Input}
\SetKwInput{Output}{Output}
    \Input{$M$: the file size of the received global model, $Bw$: the real-time network bandwidth}
    $t_{cmp}, \beta$ $\leftarrow$ one data batch training \;
    $\Tilde{t}_{com}$ = $M / Bw $ \;
    $\Tilde{t}_{cmp}$ = $t_{cmp}/ \beta$ \;
    $\Tilde{t}_{total} = \Tilde{t}_{cmp} + \Tilde{t}_{com}$ \;
    \Output{$\Tilde{t}_{total}, \Tilde{t}_{com}, \Tilde{t}_{cmp}$}
\caption{
    Local Time Update.
}
\label{alg:timeupdate}
\end{algorithm}

\textbf{Workload Scheduling.} \texttt{TimelyFL} would adjust the local epoch number $E$ and partial training ratio $\alpha$ for each client in every communication round based on the estimated $\Tilde{t}_{com, c}, \Tilde{t}_{cmp, c}$ and aggregation interval $T_k$, as the relationship shown in~\ref{eq1}.
If one's unit total time is smaller than $T_k$, then the server would try to maximize its local training utility and minimize the idle time, as to assign more than one local epoch training for the next round. Otherwise, the server would assign less amount of workload to them by decreasing the model training ratio $\alpha$, which guarantees they can finish at least one local epoch training within the report time ${t_{rpt, c}}$ and catch up the global aggregation timely. We summarized the scheduler as Algorithm~\ref{alg:workload}.

\begin{algorithm}[h]
\SetKwInput{Input}{Input}
\SetKwInput{Output}{Output}
    \Input{${T_k}$: aggregation interval time, $\langle{\Tilde{t}_{cmp}}\rangle$: unit computation time, $\langle{\Tilde{t}_{com}}\rangle$: unit communication time}
    \For{each client $c \in\mathcal{S}$ \textbf{in parallel}}{
        ${E}_{c} = \max (\lfloor (T_k - \Tilde{t}_{com, c})/\Tilde{t}_{cmp, c} \rfloor , 1)$ \;
        $\alpha_{c} = \min(T_k/(\Tilde{t}_{com, c} + \Tilde{t}_{cmp, c}), 1)$ \;
        ${t_{rpt, c}} = T_k - \Tilde{t}_{com, c} \times \alpha_{c}$ \;
    } 
    \Output{$\langle{E}\rangle, \langle{\alpha}\rangle, \langle{t_{rpt}}\rangle$}
\caption{
    Workload Scheduling.
}
\label{alg:workload}
\end{algorithm}

%% file: sections/4_exp.tex
\section {Experiment} \label{experiment}

\begin{table*}[h!]
\footnotesize
\centering
\caption{
Wall clock training time to reach target validation accuracy
on benchmark datasets (lower is better). “$>$ 200 hr” indicates the target accuracy was not reached.
}
\begin{tabular}{cccrrrr} \hline
\toprule
Dataset & Agg. function & Accuracy/Loss & TimelyFL & FedBuff & SyncFL \\ \hline 
\midrule
\multirow{4}{*}{CIFAR-10} & \multirow{2}{*}{FedAvg}
& 60\% & 5.50 $\pm 2.5\%$ hr & 7.86 $\pm 2.1\%$ hr (1.43×) & 76.81 $\pm 2.4\%$ hr (13.96×)\\ 
&& 70\% & 12.81 $\pm 1.8\%$ hr & $>$ 200 & 150.98 $\pm 1.7\%$ hr (11.78×)\\
\cmidrule(lr){2-6}
& \multirow{2}{*}{FedOpt}
& 60\% & 3.58 $\pm 2.5\%$ hr & 5.68 $\pm 2.6\%$ hr (1.59×) & 34.87 $\pm 2.3\%$ hr (9.74×)\\ 
&& 70\% & 6.46 $\pm 1.8\%$ hr & 18.73 $\pm 2.3\%$ hr (2.89×) & 58.84 $\pm 0.8\%$ hr (9.11×)\\
\hline 
\midrule
\multirow{4}{*}{Google Speech} & \multirow{2}{*}{FedAvg}
& 70\% & 22.90 $\pm 2.1\%$ hr & 42.71 $\pm 2.3\%$ hr (1.87×) & 103.07 $\pm 2.1\%$ hr (4.50×)\\ 
&& 80\% & 40.54 $\pm 1.2\%$ hr & 70.60 $\pm 2.0\%$ hr (1.74×) & 187.93 $\pm 1.2\%$ hr (4.64×)\\
\cmidrule(lr){2-6}
& \multirow{2}{*}{FedOpt}
& 70\% & 18.08 $\pm 1.1\%$ hr & 30.60 $\pm 1.7\%$ hr (1.69×) & 66.13 $\pm 1.2\%$ hr (3.66×)\\ 
&& 80\% & 31.39 $\pm 0.9\%$ hr & 53.36 $\pm 0.9\%$ hr (1.70×) & 107.38 $\pm 0.7\%$ hr (3.42×)\\
\hline 
\midrule
\multirow{4}{*}{Reddit} & \multirow{2}{*}{FedAvg}
& 7.0 (ppl)  & 9.56 {$\pm 3.1\%$} hr & 15.82 {$\pm 2.9\%$} hr (1.65×) & 23.36 {$\pm 1.5\%$} hr (2.44×)\\ 
&& 6.8 (ppl) & 17.99 {$\pm 0.7\%$} hr & $>$ 200 & 67.32 {$\pm 0.5\%$} hr (3.74×)\\
\cmidrule(lr){2-6}
& \multirow{2}{*}{FedOpt}
& 7.0 (ppl)  & 10.99 {$\pm 2.7\%$} hr & 14.09 {$\pm 2.8\%$} hr (1.28×) & 27.25 {$\pm 2.1\%$} hr (2.48×)\\ 
&& 6.8 (ppl)  & 12.86 {$\pm 0.6\%$} hr & $>$ 200 & 57.65 {$\pm 0.4\%$} hr (4.48×)\\
\hline 
\bottomrule
\end{tabular}
\label{tab:timing}
\end{table*}

\subsection{Experimental Settings}
\textbf{Datasets, Models, and Tasks.} To demonstrate \texttt{TimelyFL}'s effectiveness across tasks, we evaluate \texttt{TimelyFL} on three benchmark datasets from various categories of FL applications:
\begin{enumerate}
    \item \textbf{Image Classification.} The CIFAR-10 dataset \cite{Krizhevsky2009LearningML} consists of 60,000 colour images in 10 classes. There are 50,000 training images and 10,000 test images. To follow the realistic non-iid data in FL scenarios, we partition both datasets into 128 clusters using a Dirichlet distribution with $\alpha$ equals 0.1. We evaluate the dataset with ResNet-20 \cite{He2016DeepRL} model. 
    \item \textbf{Speech Recognition.}  The Google Command speech dataset \cite{Warden2018SpeechCA} covers 105,829 audio recordings collected from 2,618 clients. The training set includes recordings from 2,112 speakers, the validation set includes 256 speakers, and the test set includes 250 speakers. The data set is composed of 35 common words from the everyday vocabulary, such as "Yes", "No", "Up", and "Down". We evaluate the dataset with VGG11 \cite{Simonyan2015VeryDC} model for a 35-class keyword spotting task. We also evaluate the dataset with a lightweight model based on one related work \cite{Zhang2022FedAudioAF}, and the detailed data-preprocessing methods are presented in Appendix~\ref{datasets}.
    \item \textbf{Natural Language Processing.} Reddit \cite{reddit} consists of comments from 1,660,820 users in the Reddit forum. In this dataset, we filter the users with less than 20 words in total and restrict to the 30k most frequently used words, as the same settings in the previous work \cite{Lai2021FedScaleBM}. Then, we train the lightweight Albert \cite{Lan2020ALBERTAL} model for the next-word-prediction task. The performance is evaluated by the perplexity loss (ppl), which lower is better.
\end{enumerate}

\textbf{Experiment Setup.}
We use the FedML platform \cite{He2020FedMLAR, Zhang2021FederatedLF}, an open-source framework for FL, to execute our framework. On the CPU/GPU training side, to approach the real-world heterogeneous client system performance in emulation, we acquire the local computation times of deep learning models across hundreds of device types from the AI benchmark \cite{Ignatov2019AIBA} and communication times from Network Measurements on mobiles \cite{huang2011mobiperf}. These data will be assigned to the simulated devices we create in the experiment, the same as the settings in previous FL works \cite{Lai2021FedScaleBM, Lai2021OortEF, pyramidfl2022mobicom}. The distribution of heterogeneous system utility across simulated clients will be shown in the Appendix~\ref{datasets}.

\textbf{Evaluation Metrics and Baselines.}
We compare \texttt{TimelyFL} with \texttt{FedBuff} \cite{Nguyen2021FederatedLW, Huba2021PapayaPP} as the AsyncFL baseline. To demonstrate applicability of \texttt{TimelyFL}, we present the evaluation results using two aggregation function, FedAvg \cite{McMahan2017CommunicationEfficientLO} and FedOpt \cite{Reddi2021AdaptiveFO}. We evaluated the performance of \texttt{TimelyFL} and its baseline using the following three metrics:
\textit{test accuracy/loss}, \textit{time-to-accuracy}, and \textit{participation rate}. The participation rate is defined as the total number of rounds that the device contributes to the global update divided by the total communication round number. The rate is distributed in the interval between 0 and 1, which implies how often a client participates in the global model update.

\textbf{Hyperparameter Settings.}
We searched for the client learning rate in a range from $10^{-6}$ to $10^0$, server learning rate in a range from $10^{-4}$ to $10^0$, input batch size in a range from $8$ to $256$, and total training round in a range from $1000$ to $10000$. The aggregation goal and aggregation participation target is searched from 30\% to 50\% of training concurrency per round for \texttt{FedBuff} and \texttt{TimelyFL}, respectively. We list the detailed hyperparameter selection for each experiment setup in the Appendix~\ref{hyper}.

\begin{figure*}[h]
     \centering
     \begin{subfigure}[b]{0.66\columnwidth}
         \centering
         \includegraphics[width=\columnwidth]{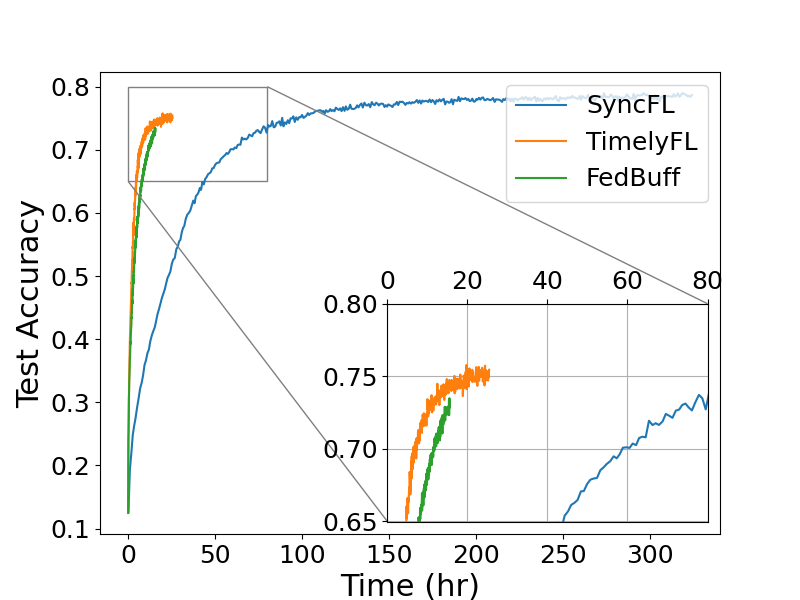}
         \caption{CIFAR-10 with FedOpt}
         \label{fig:4a}
     \end{subfigure}
     \hfill
     \begin{subfigure}[b]{0.66\columnwidth}
         \centering
         \includegraphics[width=\columnwidth]{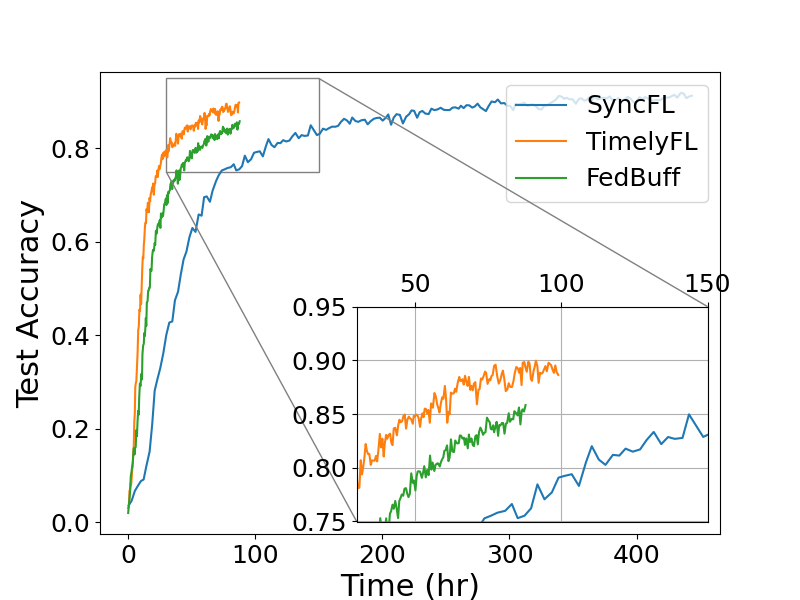}
         \caption{Google Command with FedOpt}
         \label{fig:4b}
     \end{subfigure}
     \hfill
     \begin{subfigure}[b]{0.66\columnwidth}
         \centering
         \includegraphics[width=\columnwidth]{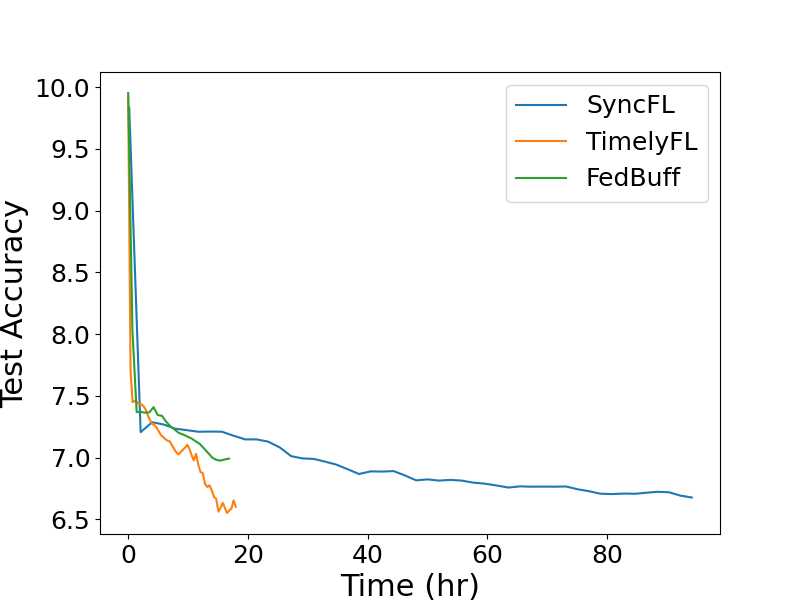}
         \caption{Reddit with FedOpt}
         \label{fig:4c}
     \end{subfigure}
        \caption{Time-to-accuracy performance for SyncFL, \texttt{FedBuff} and \texttt{TimelyFL}.}
        \label{fig:time-acc}
\end{figure*}

\subsection{End-to-End Performance} \label{performance}

We begin by comparing the end-to-end performance of \texttt{TimelyFL} on benchmark datasets, conducting on the CPU/GPU-based training. The training concurrency is set to 128 for CIFAR-10 related experiments, 20 for Google speech related experiments, and 100 for Reddit related experiments. The communication round is set to be 2000, 1000, and 500 for CIFAR-10, Google speech, and Reddit, respectively. For both \texttt{FedBuff} and \texttt{TimelyFL}, we set the aggregation goal and aggregation participation target equal to 50\% of training concurrency for a fair comparison. We run each experiment five times with different random seeds and report its mean and standard deviation for the time consumption in the Table~\ref{tab:timing}.


\begin{figure}
\centering
\begin{subfigure}[b]{0.5\textwidth}
    \centering
   \includegraphics[width=0.6\textwidth]{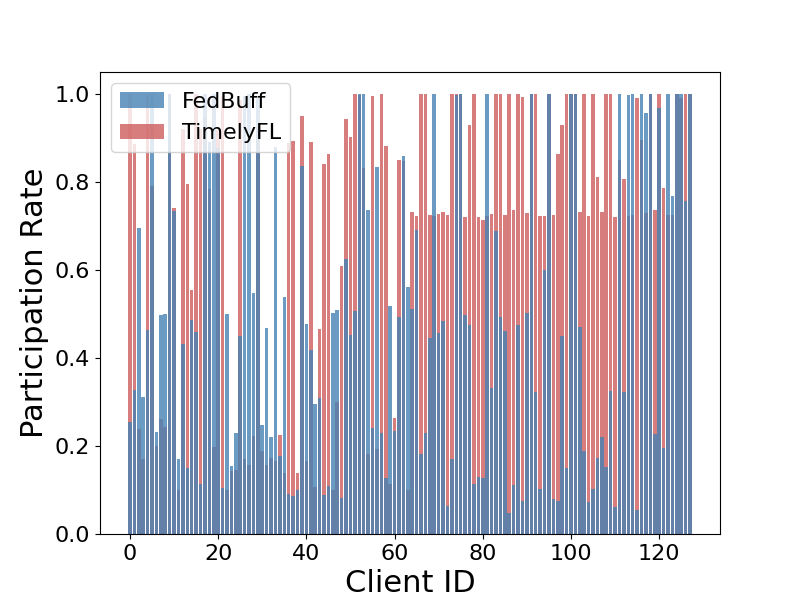}
   \caption{Participation rate for each client}
   \label{fig:Ng1} 
\end{subfigure}

\begin{subfigure}[b]{0.5\textwidth}
    \centering
   \includegraphics[width=0.6\textwidth]{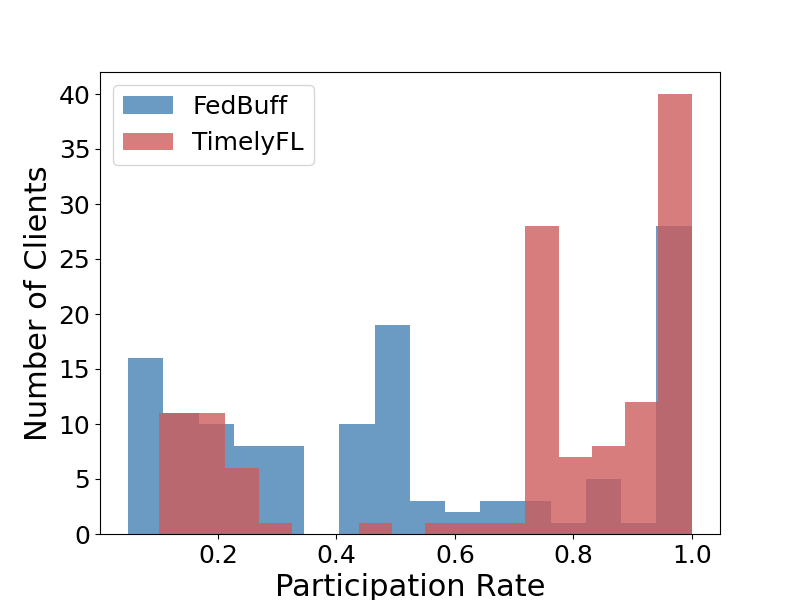}
   \caption{Participation rate distribution}
   \label{fig:Ng2}
\end{subfigure}

\caption{Participation rate evaluation.}
\label{fig:pf}
\end{figure}

\textbf{Speedup of \texttt{TimelyFL}.} Given the same heterogeneous data, \texttt{TimelyFL} achieves the shortest training time to reach all target accuracy/loss. Table~\ref{tab:timing} shows the training time needed to converge to the target accuracy/loss for each strategy considered. Compared to \texttt{TimelyFL}, synchronous FL requires 2.44 - 13.96× more times, and \texttt{FedBuff} needs 1.28 - 2.89× in terms of wall clock time. 
Besides of the time-to-accuracy speedup, \texttt{TimelyFL} also harvests test accuracy increment compared to \texttt{FedBuff} within the same communication rounds. As the learning curve in the Figure~\ref{fig:time-acc}, \texttt{TimelyFL} achieves 3.27\% and 4.01\% higher final accuracy on CIFAR-10 and Google Command, and 0.43 lower ppl on Reddit in comparison to \texttt{FedBuff} with FedOpt. Under FedAvg, \texttt{TimelyFL} achieves 4.93\% and 6.25\% higher final accuracy on CIFAR-10 and Google Command, respectively, and 0.20 lower ppl on Reddit compared to \texttt{FedBuff}.

\begin{table*}[h!]
\footnotesize
\centering
\caption{
Wall clock training time to reach target validation accuracy on benchmark datasets (lower is better).}

\begin{tabular}{cccrrrr} \hline
\toprule
Dataset & Agg. function & Accuracy & TimelyFL & FedBuff & SyncFL \\ \hline 
\midrule
\multirow{4}{*}{Google Speech} & \multirow{2}{*}{FedAvg}
& 70\% & 2.23 $\pm 2.1\%$ hr & 3.55 $\pm 1.9\%$ hr (1.59×) & 18.37 $\pm 0.6\%$ hr (8.24×)\\ 
&& 80\% & 4.16 $\pm 1.3\%$ hr & 6.13 $\pm 1.4\%$ hr (1.47×) & 32.46 $\pm 0.4\%$ hr (7.80×)\\
\cmidrule(lr){2-6}
& \multirow{2}{*}{FedOpt}
& 70\% & 0.48 $\pm 1.7\%$ hr & 1.66 $\pm 1.0\%$ hr (3.46×) & 4.61 $\pm 2.1\%$ hr (9.60×)\\ 
&& 80\% & 1.13 $\pm 1.2\%$ hr & 3.25 $\pm 0.8\%$ hr (2.88×) & 7.47 $\pm 1.1\%$ hr (6.61×)\\
\hline 
\bottomrule
\end{tabular}
\label{tab:lightspeech}
\end{table*}

\subsection{Understanding the Advantages of TimelyFL}

\textbf{\texttt{TimelyFL} improves inclusiveness\footnote{In this paper, the inclusiveness increment represents the participation rate increment in the FL training.}}. In Table~\ref{tab:timing}, we view the SyncFL as the standard baseline that does not include any asynchronous technique and \texttt{FedBuff} as the baseline that only introduces a fixed buffer size to accelerate the training. Instead of fixed buffer size, \texttt{TimelyFL} adopts a flexible buffer size controlled by aggregation interval time, which allows more available clients to participate in the global update per round. As illustrated in Figure~\ref{fig:pf}, 66.4\% of devices are able to achieve an increased participation rate, and the average participation rate per client increases by 21\% in \texttt{TimelyFL} compared to \texttt{FedBuff} under the CIFAR-10 experiment setting we implemented in the last section. 
The average participation rate increment is the main reason for the time-to-accuracy speed-up. As each client joins the global model update more rapidly, the learning efficiency increases during the FL training. In addition, combined with more devices contributing to the global model more frequently, \texttt{TimelyFL} improves inclusiveness during the model training compared to \texttt{FedBuff}.

\begin{figure}
  \begin{center}
    \includegraphics[width=0.35\textwidth]{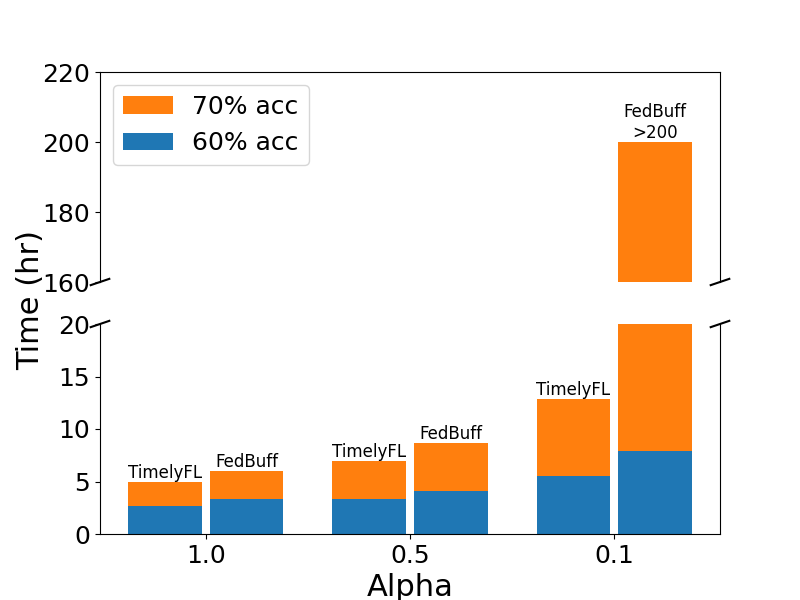}
  \end{center}
  \caption{Time-to-accuracy performance under different non-iid distribution.}
  \label{fig:alpha}
\end{figure}

The contribution of inclusiveness for model performance is especially significant when training on the non-iid dataset, where each client brings a unique local update to the global model. To demonstrate our point, we test both \texttt{TimelyFL} and \texttt{Fedbuff} with FedAvg as an aggregator on the CIFAR-10 dataset using a non-iid partition. As shown in Figure~\ref{fig:alpha}, as the parameter for Dirichlet distribution goes up, the convergence-time gap between \texttt{TimelyFL} and \texttt{Fedbuff} increases as well, which demonstrates our advantage for non-iid data training compared to \texttt{Fedbuff}.

\begin{figure}
  \begin{center}
    \includegraphics[width=0.35\textwidth]{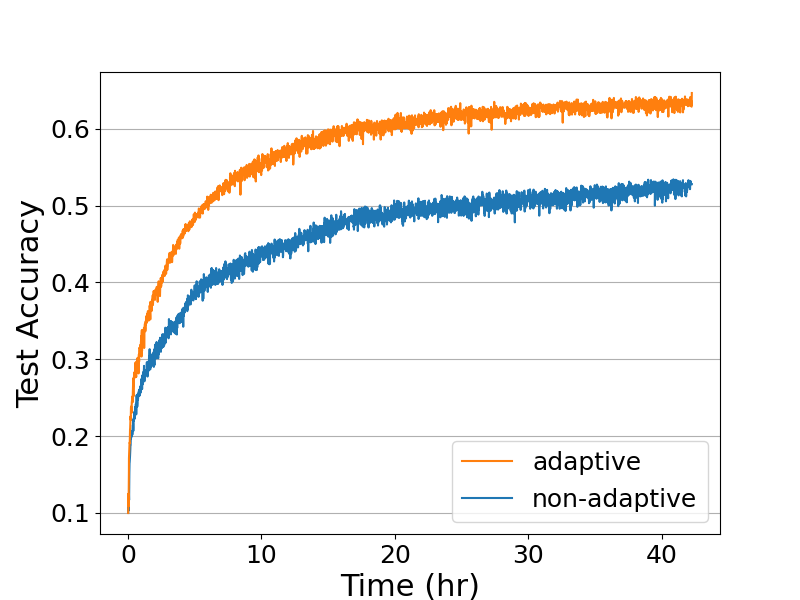}
  \end{center}
  \caption{\texttt{TimelyFL} performance under adaptive and non-adaptive workload schedule.}
  \label{fig:adapative}
\end{figure}

\textbf{\texttt{TimelyFL} is heterogeneity-aware.} Under cross-device federated learning, most participating entities have limited computing capability and intermittent connectivity. As such, it could not be guaranteed that devices would complete their training workload in every communication round as assigned initially. To effectively resist the disturbance, the training hyperparameters, such as the partial training ratio and local epoch number, should be adaptively scheduled based on the real-time capability of each device. To demonstrate our strategy, we test the training under the \texttt{TimelyFL} framework both with and without adaptive workload scheduling on the CIFAR-10 dataset, with the training concurrency equals to 64. Figure~\ref{fig:adapative} shows the learning curves for both scenarios. With adaptive workload scheduling, \texttt{TimelyFL} saves 4.09$\times$ convergence time to 50\% accuracy and 10.89\% test accuracy increment, which illustrates that real-time workload scheduling essentially improves both learning efficiency and accuracy.

\textbf{\texttt{TimelyFL} is effective on the lightweight model.} To investigate the effectiveness of the lightweight model on the \texttt{TimelyFL} framework, we implement one lightweight model on the Google Speech Commands dataset for the keyword spotting task. Following one previous work \cite{Zhang2022FedAudioAF}, we choose the model that consists of two convolution layers followed by one Gated Recurrent Units (GRU) layer. An average pooling layer is connected to the GRU output, which is then fed through two dense layers to generate the predictions. The parameter size of this model is equal to 79044.  We adopt the same baseline selections as in Section~\ref{experiment}. The hyperparameters for the experiments are listed in Section~\ref{hyper}. The experiment results are summarized in Table~\ref{tab:lightspeech}. TimelyFL achieves a higher convergence speed compared with the other two strategies before reaching the test accuracy, which confirms the simulation results elaborated in Section~\ref{performance} and demonstrates the effectiveness of the \texttt{TimelyFL} on the lightweight model architecture.


%% file: sections/5_conclusion.tex
\section {Conclusion}
In this work, we propose \texttt{TimelyFL}, a heterogeneity-aware asynchronous FL scheme with adaptive partial training.
To include more available devices joining in global aggregation in a timely manner, \texttt{TimelyFL} introduces partial model training to the slow-speed devices. Moreover, \texttt{TimelyFL} is resilient to system heterogeneity by adjusting the local training workload based on the real-time resource capabilities of each client during FL training.
Our experimental results demonstrate that \texttt{TimelyFL} could outperform major AsyncFL proposals in terms of both time-to-accuracy and test accuracy. 

%% file: sections/appendix.tex
\clearpage
\appendix
\label{sec:appendix}
\section{Appendix}
\subsection{Experiment Settings}


\subsubsection{Computing Infrastructure}
The simulation experiments are conducted on a computing server with one GPU. The server is equipped with AMD EPYC 7502 32-Core Processor and 1024G memory. The GPU is NVIDIA RTX A4000.

\subsubsection{Datasets and Models} \label{datasets}
\begin{figure}
     \centering
     \begin{subfigure}[b]{0.5\textwidth}
         \centering
         \includegraphics[width=0.6\textwidth]{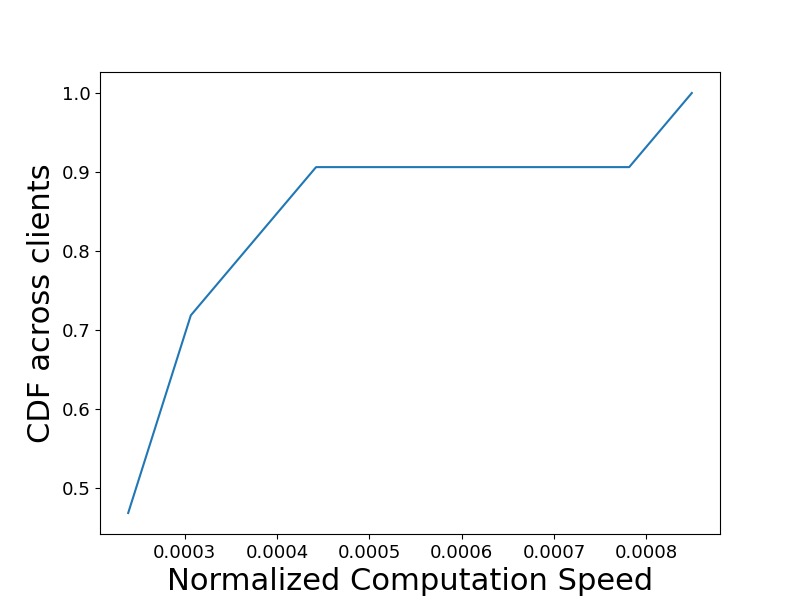}
         \caption{Diverse computation efficiency in AI Benchmark}
         \label{fig:9a}
     \end{subfigure}
     \hfill
     \begin{subfigure}[b]{0.5\textwidth}
         \centering
         \includegraphics[width=0.6\textwidth]{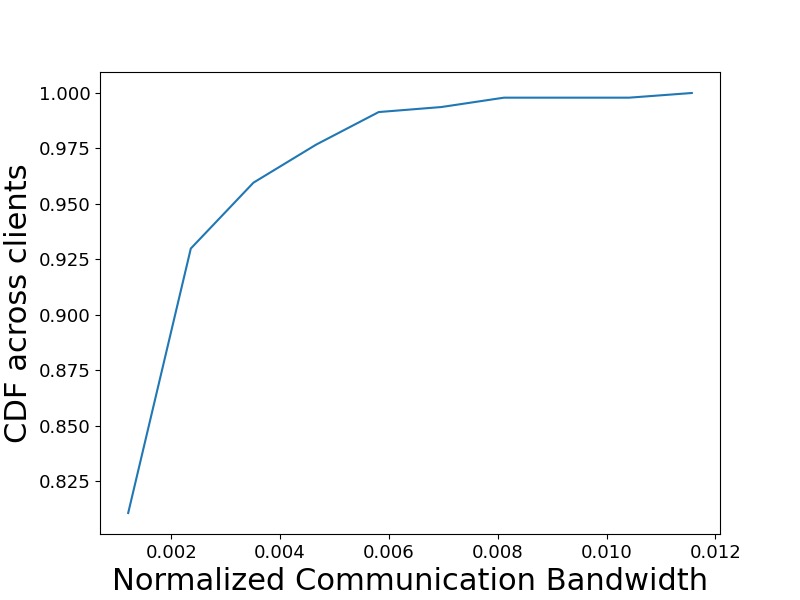}
         \caption{Diverse communicate efficiency in Mobiperf}
         \label{fig:9b}
     \end{subfigure}
        \caption{Heterogeneous system utility across simulated clients.}
        \label{fig:9}
\end{figure}

\textbf{AI Benchmark.} AI Benchmark \cite{Ignatov2019AIBA} is a public dataset that is designed for evaluating the performance of important AI tasks on mobile devices. AI Benchmark provides diverse models' training and inference speed across various devices, including chipsets from Qualcomm, HiSilicon, Samsung, MediaTek, and Unisoc.  Figure~\ref{fig:9a} illustrates the distribution of the computation efficiency across clients in the AI Benchmark. The slowest device would take around 13.3$\times$ computational times than the fastest device for the same task. To approach the dynamic availability of devices, such as low-power mode or multi-process running, we design a coefficient $w$ as follows:
\begin{equation}
  \begin{gathered}
  x \sim \mathcal{N}(1,\,0.3) \\\
  w =
    \begin{cases}
      1 & \text{$x \leq 1$}\\
      x & \text{$1 \leq x \leq 1.3$}\\
      1.3 & \text{$x \ge 1.3$}
    \end{cases}
  \end{gathered}
\label{w}
\end{equation}
In this work, we assign the values from AI Benchmark as base computation time to the clients to emulate real devices, analogous to the usage in FedScale \cite{Lai2021FedScaleBM}. We also generate the coefficient $w$ every round for each client to simulate the natural disturbance to availability. The local computation time in each round equals the product of $w$ and the base computation time for each client.

\textbf{MobiPerf.} MobiPerf is a public dataset for measuring network performance on mobile devices, which collects the available
cloud-to-edge network throughput of over 100k worldwide mobile clients.
Figure~\ref{fig:9b} illustrates the distribution of communication consumption of MobiPerf. Note that the best communication channel can be
200$\times$ better than the worst one. We randomly assign a value from MobiPerf to a simulated device every communication round to emulate intermittent connectivity in a real deployment.

\textbf{CIFAR-10.} The CIFAR-10 dataset \cite{Krizhevsky2009LearningML} consists of 60,000 32x32 colour images in 10 classes. There are 50,000 training images and 10,000 test images. We normalize the images by the mean and standard deviation of the dataset. We evaluate the dataset with ResNet-20 \cite{He2016DeepRL} model. To emulate the realistic non-iid distribution, we partition the dataset using a Dirichlet distribution, following the previous works \cite{Nguyen2021FederatedLW}.

\textbf{Google Command.} The Google Command speech dataset \cite{Warden2018SpeechCA} covers 105,829 audio recordings collected from 2,618 clients. The training set includes recordings from 2,112D speakers, the validation set includes 256 speakers, and the test set includes 250 speakers. The data set is composed of 35 common words from the everyday vocabulary, such as "Yes", "No", "Up", and "Down". We evaluate the dataset with VGG11 \cite{Simonyan2015VeryDC} model and a lightweight model based on one related work \cite{Zhang2022FedAudioAF} for a 35-class keyword spotting task.

For the VGG11-based experiment on Google Speech Commands, we use the Mel-frequency cepstral coefficients (MFCC) method to pre-process the raw audio data. Specifically, a sequence of overlapping Hamming windows is applied to the raw speech signal with a time shift of 10 ms and window size of 25ms. The MFCC is used for training the keyword spotting model.

For the lightweight model experiment, to pre-process the raw audio data, a sequence of overlapping Hamming windows is applied to the raw speech signal with a time shift of 10 ms. We calculate the discrete Fourier transform (DFT) with a frame length of 1,024 and compute the Mel-spectrogram with a dimension of 128. The Mel-spectrogram is used for training the keyword spotting model. We follow \cite{Zhang2022FedAudioAF} for this setup.

\textbf{Reddit.} Reddit \cite{reddit} consists of comments from 1,660,820 users in the Reddit forum. Each client corresponds to a user, whose
data are all of their personal posts. Thus it follows the real non-iid data under FL scenarios. In this dataset, we filter the users with less than 20 words in total and restrict to the 30k most frequently used words, as the same settings in the previous work \cite{Lai2021FedScaleBM}. Then, we train the lightweight Albert \cite{Lan2020ALBERTAL} model for the next-word-prediction task. The performance is evaluated by the perplexity loss (ppl), which lower is better.

\subsubsection{Hyperparameter Settings} \label{hyper}
We searched for the client learning rate in a range from $10^{-6}$ to $10^0$, server learning rate in a range from $10^{-4}$ to $10^0$, input batch size in a range from $8$ to $256$, and total
training round in a range from $1000$ to $10000$. The aggregation goal and aggregation participation target is searched from 30\% to 50\% of training concurrency per round for \texttt{FedBuff} and \texttt{TimelyFL}, respectively.

After hyper-parameter searching, we fixed the following hyperparameters: for CIFAR-10 related experiments, the total training round is 2000, and training concurrency is 128 for all setups. The aggregation goal and aggregation participation target is 50\% of the training concurrency for both \texttt{FedBuff} and \texttt{TimelyFL}. For CIFAR-10 with FedAvg related experiments, the batch size is 8, and the client learning rate is 0.8. For CIFAR-10 with FedOpt related experiments, the batch size is 10, the client learning rate is 0.03, and the server learning rate is 0.001 with ADAM as server optimizer. 

For Google command related experiments with VGG11 model, the total training round is 1000, and training concurrency is 20 for all setups. The aggregation goal and aggregation participation target is 50\% of the training concurrency for both \texttt{FedBuff} and \texttt{TimelyFL}.
The batch size is 32, and the client learning rate is 0.01. Under the FedOpt, the server learning rate is 0.001 with ADAM as server optimizer.

For Google command related experiments with the lightweight model, the total training round is 5000, and training concurrency is 106 for all setups. The aggregation goal and aggregation participation target is 50\% of the training concurrency for both \texttt{FedBuff} and \texttt{TimelyFL}.
The batch size is 16, and the client learning rate is 0.1 under the FedAvg. Under the FedOpt, the client learning rate is 0.05 for synchrounous FL and \texttt{TimelyFL}, and the client learning rate is 0.2 for \texttt{FedBuff}. The server learning rate is 0.001 with ADAM as server optimizer for all setups.

Finally, for Reddit related experiments, the total training round is 500, and training concurrency is 20 for all setups. The aggregation goal and aggregation participation target is 50\% of the training concurrency for both \texttt{FedBuff} and \texttt{TimelyFL}. The batch size is 20, and the client learning rate is 0.0005 for SyncFL and \texttt{TimelyFL}, and 0.0003 for \texttt{FedBuff}. Under the FedOpt, the server learning rate is 0.001 with ADAM as server optimizer.

\subsection{System Performance}

\subsubsection{Partial Training Performance} \label{partial_experiment}
\begin{figure}
  \begin{center}
    \includegraphics[width=0.4\textwidth]{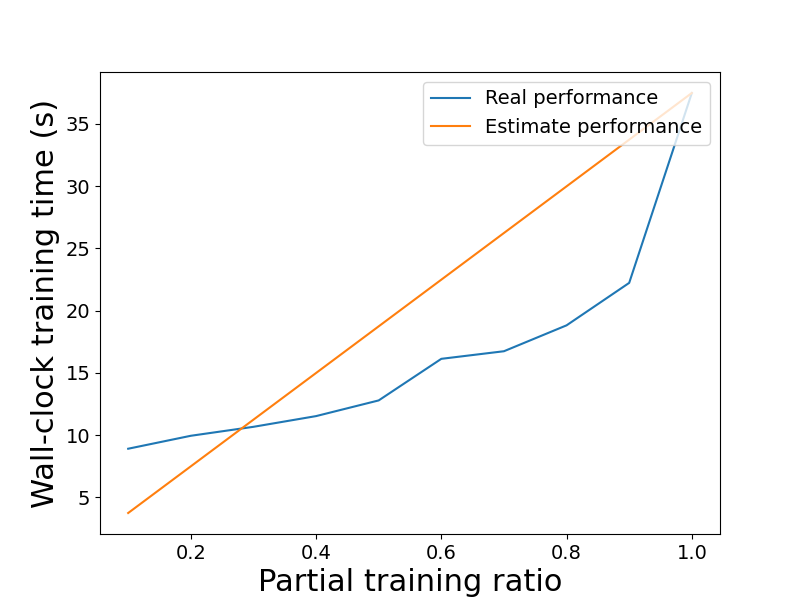}
  \end{center}
  \caption{Partial training system performance in on real edge devices.}
  \label{fig:10b}
\end{figure}

Due to different parameters and tensor shapes among different layers, the training time (computational time of the forward and backward propagation) is not strictly linear to the trainable layer numbers and varies with the model structures. For simplicity and generality, we define the training time of the partial model as the linear multiplication of the training time of the full model and the training ratio. This linear relationship is verified through our real measurement on a Samsung Galaxy S20 with ResNet-20 model using MNN \cite{Jiang2020MNNAU} library. As shown in Figure~\ref{fig:10b}, most of the test results are below the linear straight line (except the ratio is below 0.2), justifying the rationality of our choice. 



